\pdfoutput=1

\documentclass[11pt]{article}

\usepackage[review]{EMNLP2023}

\usepackage{times}
\usepackage{latexsym}

\usepackage[T1]{fontenc}

\usepackage[utf8]{inputenc}

\usepackage{microtype}

\usepackage{inconsolata}

\usepackage{subcaption}
\usepackage{booktabs}
\usepackage{makecell}
\usepackage{svg}
\usepackage{amsmath}
\usepackage{placeins}
\usepackage{multirow}
\usepackage{hyperref}
\hypersetup{pdfborder={1 1 0}}
\usepackage{cleveref}
\definecolor{ORANGE}{rgb}{0.90, 0.69, 0.12}
\definecolor{BLUE}{rgb}{0.47, 0.61, 0.89}
\definecolor{GREEN}{rgb}{0.56, 0.70, 0.59}
\definecolor{url}{rgb}{0,0,0.93}
\usepackage{tikz} 
\usepackage{lipsum}
\usepackage{algorithm}
\usepackage{algorithmic}
\newcommand{\mybox}[2]{
    \begin{figure}[h]
        \centering
    \begin{tikzpicture}
        \node[anchor=text,text width=\columnwidth-0.1cm, draw, rounded corners, line width=0.1pt, fill=#1, inner sep=1mm] (small) {#2};
    \end{tikzpicture}
    \end{figure}
}

\usepackage{color,soul}
\usepackage{xcolor}
\usepackage{fontawesome}

\title{Large Language Models for \\Automated Open-domain Scientific Hypotheses Discovery}

\author{
    Zonglin Yang\textsuperscript{\rm 1},
    Xinya Du\textsuperscript{\rm 2},
    Junxian Li\textsuperscript{\rm 1},
    Jie Zheng\textsuperscript{\rm 3},
    Soujanya Poria\textsuperscript{\rm 4},
    Erik Cambria\textsuperscript{\rm 1}\\
    \textsuperscript{\rm 1} {\small Nanyang Technological University}
    \textsuperscript{\rm 2} {\small University of Texas at Dallas}\\
    \textsuperscript{\rm 3} {\small Huazhong University of Science and Technology}
    \textsuperscript{\rm 4} {\small Singapore University of Technology and Design}\\
    {\tt \small \{zonglin001,junxian001,cambria\}@ntu.edu.sg, xinya.du@utdallas.edu} \\
    {\tt \small jie.jay.zheng@gmail.com, sporia@sutd.edu.sg}
  }

\begin{document}
\maketitle
\begin{abstract}
Hypothetical induction is recognized as the main reasoning type when scientists make observations about the world and try to propose hypotheses to explain those observations. Past research on hypothetical induction is under a constrained setting: (1) the observation annotations in the dataset are carefully  manually handpicked sentences~(resulting in a close-domain setting); and (2) the ground truth hypotheses are mostly commonsense knowledge, making the task less challenging. In this work, we tackle these problems by proposing the first dataset for social science academic hypotheses discovery, with the final goal to create systems that automatically generate valid, novel, and helpful scientific hypotheses, given only a pile of raw web corpus. Unlike previous settings, the new dataset requires (1) using open-domain data~(raw web corpus) as observations; and (2) proposing hypotheses even new to humanity. A multi-module framework is developed for the task, including three different feedback mechanisms to boost performance, which exhibits superior performance in terms of both GPT-4 based and expert-based evaluation.
To the best of our knowledge, this is the first work showing that LLMs are able to generate novel~(``not existing in literature'') and valid~(``reflecting reality'') scientific hypotheses\footnote{Dataset, code, and generated hypotheses are available at 
\url{https://github.com/ZonglinY/MOOSE.git}.}.
\end{abstract}

\section{Introduction}
Logical reasoning is central to human cognition~\citep{goel2017reasoning}.
It is widely recognized as consisting of three components, which are deductive, inductive, and abductive reasoning~\citep{DBLP:journals/corr/abs-2303-12023}.
Hypothetical induction is considered to be an important sub-type of inductive reasoning~\citep{norton2003little}.
It is recognized as the main reasoning type when scientists make observations about the world and try to propose hypotheses to explain the observations. 

For example, the proposal of Geocentrism, Heliocentrism, and Newton's law of universal gravitation based on the observations of the motion of (celestial) objects can be seen as a result of hypothetical induction.
Hypothetical induction is a process of knowledge exploration from observations to hypotheses: it is challenging because it involves the exploration of knowledge that is even new to humanity.
\begin{figure}[t]
\centering
\resizebox{1.0\columnwidth}{!}{
\includegraphics[]{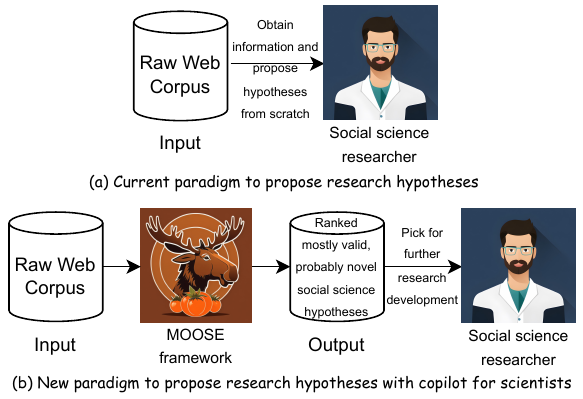}
}
\caption{Comparison of the two paradigms for scientific hypotheses formulation.
The new paradigm shows the role of the MOOSE framework~(scientist's copilot) and the new task setting of hypothetical induction.}
\label{fig:intro_figure}
\end{figure}
Recent research on this has two main limitations~\citep{yang-etal-2024-language}. 
Firstly, the observations in their dataset have already been manually selected from the raw web corpus, resulting in a close-domain setting.
As a result, a developed system for this dataset relies on already manually selected observations, and cannot utilize the vast raw web corpus to propose hypotheses.
Secondly, the ground truth hypotheses are mostly commonsense knowledge~(e.g., Newton's law), making the task less challenging since LLMs might have already seen them during pretraining.
To this end, we propose a new task setting of hypothetical induction, which is to generate novel and valid research hypotheses targeting being helpful to researchers while only given (vast) raw web corpus (Figure~\ref{fig:intro_figure}).

This hypothesis formation process is seen as the first step for scientific discovery~\citep{wang2023scientific}.
We call this task as ``au\textbf{TO}mated open-do\textbf{MA}in hypo\textbf{T}hetical inducti\textbf{O}n~(TOMATO)''. 
It is ``automated'' since a method for this task should automatically propose hypotheses with few human efforts;
It is open-domain since it is not restricted by any manually collected data.

For the TOMATO task, we constructed a dataset consisting of 50 recent social science papers published after January 2023 in top social science journals.
For each paper, social science experts collect its main hypothesis, identify its background and inspirations, find semantically similar contents for its background and inspirations from the web corpus, collect the full passage for each matched content, and use all collected web passages as raw web corpus. 
Although the new dataset involves many manual selection processes, the manually selected contents are used more as benchmarking human performance for comparison.
In the TOMATO task, a method is required to only utilize the raw web corpus in the dataset to propose hypotheses.
In addition, the raw web corpus is mostly from common news, Wikipedia, and business reviews, which means it can easily expand in scale without much human involvement.

To tackle the TOMATO task, we develop a multi-module framework called MOOSE based on large language model (LLM) prompting (Figure~\ref{fig:tomato}).
To further improve the quality of the generated hypotheses, we also propose three different feedback mechanisms~(present-feedback, past-feedback, and future-feedback) to use LLMs to retrospect and improve the LLM-generated hypotheses for better quality.
For present-feedback, the intuition is that, for some modules, their generation can be evaluated by other LLMs and be provided with feedback, which can be utilized by the modules to refine their generation by taking the feedback and previous generation as input and generating again. 
Some modules can have feedback instantly after their generation to improve themselves.
But just like the reward mechanism in reinforcement learning, some rewards~(feedback) might be hard to obtain instantly, but need to wait for feedback for a future module.
Similarly, we develop past-feedback where a module can benefit from the feedback for a future module.
The last one is future-feedback, where a current module can provide justifications for the current module's generation to help a future module's generation, or can provide some initial suggestions which a future module can build upon to further provide more in-depth generation.

Both GPT-4~\citep{DBLP:journals/corr/abs-2303-08774} evaluation and expert evaluation indicate that MOOSE performs better than an LLM~\citep{DBLP:conf/nips/Ouyang0JAWMZASR22} based baseline, and each of the three feedback mechanisms can progressively improve the base framework.
During expert evaluation, many hypotheses generated by MOOSE are recognized by social science researchers to be both novel~(``not existing in the literature'') and valid~(``reflecting reality'').
To the best of our knowledge, this is the first work showing that LLMs can be leveraged to generate novel and valid research hypotheses, 
indicating the potential for LLMs to serve as a ``copilot'' for scientists.

\section{Related Work}

\subsection{NLP Methods for Scientific Discovery}
\citet{DBLP:journals/corr/abs-2302-14233} propose a dataset where each data consists of a research goal, a corpus pair, and a
discovery.
However, (1) their task needs a human-provided research goal and a pairwise corpus for discovery, which is not an automated setting and has a limited application scope; (2) the
discovery is not from recent publications.
\citet{DBLP:journals/corr/abs-2305-14259} is a concurrent work of ours. Compared the first version of two papers, they do not have an iterative feedback for novelty, reality, and clarity. Later they add for novelty, but still lack the other two.
These aspects are required by inductive reasoning, and there's an implicit trade-off between reality and novelty. Only stressing on novelty might lead to incorrect and vague generation.
\citet{bran2023chemcrow} focuses on integrating computational tools in the chemistry domain, but not on providing novel chemistry findings or hypotheses.
\citet{DBLP:journals/corr/abs-2304-05332} focuses on using LLMs to design, plan, and execution of scientific experiments, but not on finding novel hypotheses.

\subsection{LLM-based Self Feedback}
Self-refine~\citep{DBLP:journals/corr/abs-2303-17651} investigates feedback but it only focuses on present-feedback~(our framework also proposes past-feedback and future-feedback), and it is not specially designed for inductive reasoning tasks.
Other similar works to self-refine~\citep{DBLP:journals/corr/abs-2210-03350,DBLP:journals/corr/abs-2302-12813,DBLP:conf/emnlp/YangTPK22,DBLP:journals/corr/abs-2303-11366} also only focus on present-feedback, and their feedback is not multi-aspect nor iterative compared to ours.

Our present-feedback is developed upon a multi-aspect over-generate-then-filter mechanism~\citep{yang-etal-2024-language}. 
However, they only utilize LLMs to ``filter'' but not to provide feedback.

\section{Dataset Collection}
\label{sec:dataset_collection}
\begin{figure}[t]
\centering
\resizebox{0.9\columnwidth}{!}{
\includegraphics[]{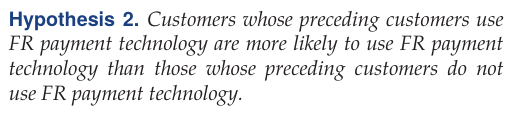}
}
\caption{A selected hypothesis in a social science publication collected in our dataset.}
\label{fig:hypothesis}
\end{figure}
\begin{figure}
\centering
\begin{subfigure}[b]{0.45\columnwidth}
   \includegraphics[width=1\linewidth]{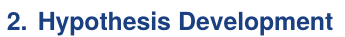}
   \label{fig:hypothetica_development_section} 
\end{subfigure}
\begin{subfigure}[b]{0.30\columnwidth}
   \includegraphics[width=1\linewidth]{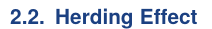}
   \label{fig:herding_effect}
\end{subfigure}
\caption{Hypothetical development section and a particular theory subsection for developing hypotheses.}
\label{fig:inspiration}
\end{figure}


\begin{table}[]
\centering
\resizebox{0.65\columnwidth}{!}{
\begin{tabular}{c|lc}
\toprule
\multirow{7}{*}{Social Science} & Communication             & 5  \\
                                & Psychology                & 7  \\ 
                                & Human Resource Management & 8  \\
                                & Information System        & 8  \\
                                & International Business    & 5  \\
                                & Management                & 6  \\
                                & Marketing                 & 11 \\ 
\bottomrule
\end{tabular}
}
\caption{Statistics of subject distribution of the dataset.}
\label{tab:dataset_subject}
\end{table}
\begin{figure*}[t]
\centering
\resizebox{2.0\columnwidth}{!}{
\includegraphics[]{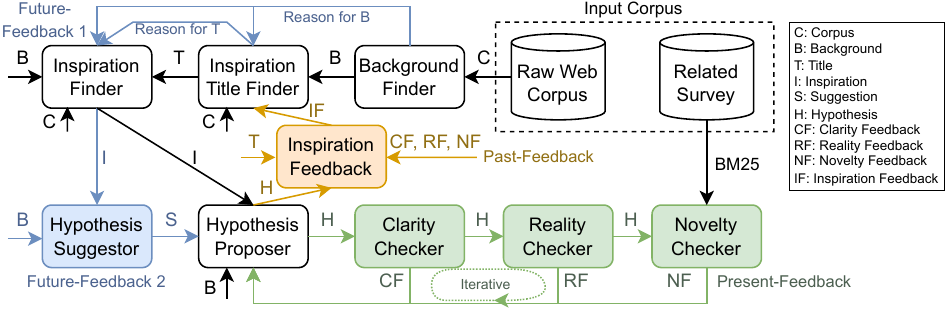}
}
\caption{MOOSE: Our multi-module framework for TOMATO task. The black part is the base framework; \textcolor{ORANGE}{orange part} represents past-feedback.; \textcolor{GREEN}{green part} represents present-feedback; \textcolor{BLUE}{blue part} represents future-feedback.
Each capitalized letter represents the generation of one of the modules. 
The same capitalized letter represents the same regardless of its color.
If a module has an input arrow pointing in with a capitalized letter, it represents that this module utilizes one of its previous modules' generation~(which has the same letter pointing out) as input.
}
\label{fig:tomato}
\end{figure*}

In this section, we take one publication~\citep{gao2023improving} in our dataset as an example to illustrate the dataset collection process. 
In total, there are 50 papers published after January 2023. 
Table~\ref{tab:dataset_subject} shows the statistics of the subject distribution.

Most social science publications highlight their hypotheses. 
Figure~\ref{fig:hypothesis} shows our selected main hypothesis in the example publication.
The research backgrounds are given in the introduction section.
In this example paper, the background is about facial recognition payment technology's usage in society.
Most social science publications also have a ``Hypothesis Development'' section~(some may call it by other names, e.g., ``Theoretical Development'').
For example, the left part (``Hypothesis Development'') in Figure~\ref{fig:inspiration} shows the title of this section in the example paper.
In this section, several theories used to develop the main hypothesis are separately introduced.
Usually, each theory takes one subsection.
For example, the right part (``Herding Effect'') in Figure~\ref{fig:inspiration} shows the title of a subsection, which is a particular theory being used as an inspiration, which with the background can develop the hypothesis in Figure~\ref{fig:hypothesis}.

For each publication in our dataset, we identify its main hypothesis, research background, and inspirations, where the background and inspirations together provide enough information to be possible to develop the hypothesis. 
We also abstract the reasoning process from background and inspirations to hypothesis and note it down for each publication in our dataset.
In this selected example, the reasoning process is easy, but it has medium difficulty for researchers to associate the inspiration~(herding effect) to the background.
For each publication, we include an expert-evaluated complexity for both the reasoning process and the association of the inspiration to the background~(details in \S\ref{appen:dataset_conplexity}). 

Instead of directly copying the background and inspirations from the paper to construct the dataset, we try to find semantically similar text contents from the web corpus as a substitution to avoid data contamination and fit the requirement of TOMATO task that a system should propose novel and valid research hypotheses only given raw web corpus.
In the example paper, we find news sentences reporting the usage of facial recognition payment as ground truth background and a Wikipedia description of the herding effect as ground truth inspiration.
We also collect the web link and the full text of the manually selected web passages for backgrounds and inspirations to be used as raw web corpus.

In addition, we collect the link and the publication date for all fifty papers. 
We also collected fourteen survey papers in related fields that might help check the novelty of the hypotheses.
The dataset is fully constructed by a social science PhD student.
We illustrate why the dataset shouldn't be collected by automatic methods in \S\ref{appen:why_cant_automatic_collect_dataset}.


\section{Methodology}
\label{sec:methodology}

In general, our method consists of a base multi-module framework and three feedback mechanisms~(past-feedback, present-feedback, and future-feedback).
We call the full framework as \textbf{M}ulti-m\textbf{O}dule framew\textbf{O}rk with pa\textbf{S}t present future fe\textbf{E}dback~(MOOSE). 
The base framework without any feedback is called MOOSE-base.
MOOSE is described in Figure~\ref{fig:tomato} and Algorithm~\ref{alg:algorithm}. 

\subsection{Base Framework}

The base framework is developed based on the intuitive understanding of how social science researchers propose an initial research hypothesis.

Firstly, a researcher needs to find a suitable research background, e.g., facial recognition payment system's impact.
This background should be proposed with a deep understanding of the societal world.
Accordingly, we develop a background finder module, which reads through raw web corpus to find reasonable research backgrounds.

Secondly, since the proposed hypothesis should be novel, directly copying from raw web corpus usually is not enough.
A good social science hypothesis should contain an independent variable and a dependent variable, and describe how the independent variable can influence the dependent variable. 
Therefore, building connections between two variables that have not been known for established connections contributes to a novel hypothesis.
We hypothesize that proper inspiration can help this connection-building process, since 
it
might serve as one of the variables itself, or might help to find such variables.
However, it could consume lots of computing resources and even be practically impossible if the framework searches over the full web corpus for every found background. 
Nevertheless, it could be much more viable if only searching over the titles of the corpus, and then only finding inspirational sentences in the passages which match the selected titles.
Accordingly, we develop an inspiration title finder module and an inspiration finder module, together to find proper inspirations given a background.

Lastly, a hypothesis proposer module can utilize backgrounds and inspirations for hypotheses.

In general, MOOSE-base consists of a list of serializable generation modules $M_0, M_1, ..., M_n$ that function sequentially.
The input of a module $M_i$ is from the output of previous modules $M_{j, j<i}$ and a raw web corpus $C$~(and optionally a related survey corpus).
$M_i$'s output is represented as $o_i$.
Feedback to $o_i$ is represented as $f_i$.

\subsection{Present-Feedback}
LLMs are not perfect and can lead to flaws in the generation, especially for those modules that undertake a difficult task.
Previous work on hypothetical induction~\citep{yang-etal-2024-language} tackles this problem by 
leveraging LLMs to identify flaws in the generation and filters those with huge flaws. 
Here we take a step further that instead of filtering, LLMs are leveraged to provide feedback, so that a generation can be improved rather than just filtered.

Accordingly, we define \emph{present-feedback} as when an output $o_i$ can be directly evaluated and provided feedback $f_i$~(by LLMs or experts, here we use LLMs) in terms of some aspects, $o_i$ and $f_i$ are used as additional inputs to $M_i$, so that $M_i$ can regenerate $o_i$ to refine the previous one with $f_i$.

We implement present-feedback on the \emph{Hypotheses Proposer} module, since it is a key module that undertakes a very difficult task.
In terms of what aspects should the feedback focus on, \citet{yang-etal-2024-language} propose four aspects according to the philosophical definition and requirement for hypothetical induction~\citep{norton2003little}.
The aspects are whether the hypothesis
(1) is consistent with observations;
(2) reflects reality;
(3) generalizes over the observations;
(4) is clear and meaningful.

In MOOSE, we basically adopt the four aspects but reframe them to better fit the current task.
Specifically, aspect~(2) contains aspect~(1) most of the time~(unless the observations are wrongly described). 
To save computing power, we adopt aspect~(2) but not aspect~(1).
In addition, we reframe aspect~(3) as whether the hypothesis is novel, and reframe aspect~(4) as whether the hypothesis is clear and provides enough details.
Accordingly, we develop a reality checker module, a novelty checker module, and a clarity checker module in Figure~\ref{fig:tomato}.

\subsection{Past-Feedback}

Just like the reward mechanism in reinforcement learning, some modules' generation can only be evaluated at a future time point. 
For instance, it is hard to give feedback on the selected inspirations unless we know what hypotheses these inspirations could lead to. 
Accordingly, we develop \emph{past-feedback} as when it is hard to directly evaluate $o_i$, the framework continues to run until generating $o_{j, j>i}$, where $o_j$ is highly influenced by $o_i$ and can be directly evaluated to obtain present-feedback $f_j$.
Then $o_i$, $o_j$, and $f_j$ are utilized, possibly by an additional module implemented with an LLM, to provide past-feedback $f_i$ to $M_i$, so that $M_i$ can regenerate $o_i$ with $f_i$ to refine the previous $o_i$.

We implement past-feedback on the \emph{Inspiration Title Finder} module.
The intuition is that improper inspirations can lead to low-quality hypotheses, and it is hard to directly evaluate inspirations.

\subsection{Future-Feedback}
We also develop \emph{future-feedback}, targeting at providing additional useful information for a future module $M_j$ to generate $o_j$ in better quality.
Specifically, we develop future-feedback-1~(FF1) and future-feedback-2~(FF2).
FF1 is that in addition to $o_i$, justifications~(reasons) of $o_i$ are also provided to $M_{j, j>i}$ so that $M_j$ can better leverage $o_i$; 
FF2 is that for a key module $M_j$ that handles a very complex task, an additional module $M_{j-0.5}$ is being placed before $M_j$, so that $M_{j-0.5}$ can undertake some of the reasoning burdens of $M_j$ to improve the quality of $o_j$. 
For example, in MOOSE, $M_{j-0.5}$ is to provide preliminary suggestions for $M_j$.

Specifically in the MOOSE framework, for FF1, no additional modules are needed. 
Instead, we modify the prompt to require $M_i$ to not only generate $o_i$ but also provide the justification of $o_i$.
We implement it on the \emph{Background Finder} and the \emph{Inspiration Title Finder} modules.
The intuition is that it could be helpful if the \emph{Inspiration Title Finder} module knows not only the background but also what possible research topics could be conducted for this background so as to select suitable titles; 
it could be also helpful for the \emph{Inspiration Finder} module to know why this background was selected and what potentially helpful inspirations could be found from the passage with the corresponding selected titles.
For FF2, we implement it on the \emph{Hypothesis Proposer} module, since proposing hypotheses is a very important and complex task.
Accordingly, we develop a \emph{Hypothesis Suggestor} module~(as $M_{j-0.5}$) to provide some initial suggestions on how to utilize the inspirations and background first, and then \emph{Hypothesis Proposer}~(as $M_j$) can build upon the suggestions to generate more novel and more complicated hypotheses.

\section{Experiments}
\label{sec:experiment}

\subsection{Evaluation Metrics \& Details}
\label{subsec:metrics}
We conduct both automatic evaluation and human evaluation for the experiments.

For automatic evaluation, we adopt validness, novelty, and helpfulness as three aspects for GPT-4 to evaluate.
We choose validness and novelty because they are the two basic requirements for hypothetical induction illustrated in philosophical literature~\citep{norton2003little,yang-etal-2024-language}.
In addition, these two scores also highly resemble the current ACL review form, which requires reviewers to score submitted papers on soundness and excitement aspects.
We choose helpfulness because the final goal of the TOMATO task is to provide help and assistance for human scientists.

In \S\ref{appen:why_not_other_eval_metrics} we illustrate why we don't adopt evaluation metrics such as (1) relevance and significance, and (2) BLEU~\cite{papineni2002bleu}, ROUGE~\cite{lin2004rouge}, or METEOR~\citep{banerjee-lavie-2005-meteor}.

For human~(expert) evaluation, evaluation metrics are the same.
Three experts~(social science PhD students) take charge of the expert evaluation. 
They evaluate on 400 randomly selected hypotheses from the baseline and variants of the MOOSE framework.
To avoid any bias, they are not told which methods we are comparing; the order of generated hypotheses to compare is also randomized.
We introduce how the 400 hypotheses are selected in \S\ref{appen:how_to_select_400_hypotheses}, and the high expert agreement in \S\ref{appen:expert_qualification_and_agreement}.

Each metric is on a 5-point Likert scale.
Both experts and GPT-4 are given the same description of the scale and evaluation standard of the three aspects~(listed in \S\ref{appen:eval_metrics_description}).

Out of the metrics, we consider the novelty metric to be relatively more important than the validness metric.
Because the goal of the TOMATO task is to assist human researchers, but not to directly add the machine-proposed hypotheses to the literature.
If the hypotheses are fully valid but not novel, then they are not helpful at all; but if the hypotheses are novel but not valid, then they can still be possible to inspire human researchers to develop novel and valid hypotheses.
Helpfulness is also an important metric since it could be seen as an overall evaluation of a hypothesis.

In \S\ref{appen:consistency_expert_gpt4}, we introduce the surprisingly high consistency between expert evaluation and GPT4 evaluation, indicating that GPT-4 might be able to provide a relatively reliable evaluation for machine-generated 
social science
hypotheses.

\subsection{Baselines \& Base Model Selection}

Since the TOMATO task is to propose hypotheses given only corpus, a natural baseline is to use a corpus chunk as input, and directly output hypotheses.

Except for \S\ref{sec:base_model_selection}, we use \texttt{gpt-3.5-turbo} for each module in MOOSE.
To be fair, the baseline is also instantiated with \texttt{gpt-3.5-turbo}. 
The training data of the model checkpoint is up to September 2021, while all papers in our dataset are published after January 2023, so the model has not seen any of the collected papers in the dataset.
In \S\ref{sec:base_model_selection}, we investigate the effect of base model selection by using \texttt{Claude3-Opus}~\citep{anthropic2024claude} for each module in MOOSE.

\subsection{Main Results}
\label{subsec:main_results}
\begin{table}[]
\centering
\resizebox{1.0\columnwidth}{!}{
\begin{tabular}{l|ccc}
\toprule
                            & Validness      & Novelty        & Helpfulness    \\ \midrule
Baseline                    & 3.954          & 2.483          & 3.489          \\ \midrule
MOOSE-base                  & 3.907          & 3.081          & 3.859          \\
\quad w/ \emph{future-feedback}          & \textbf{3.955} & 3.226          & \textbf{3.953} \\
\quad w/ \emph{future-} and \emph{past-feedback} & 3.916          & \textbf{3.390} & 3.931          \\ \bottomrule
\end{tabular}}
\caption{Effect of MOOSE-base, \emph{future-feedback} and \emph{past-feedback}~(evaluated by \texttt{GPT-4}). MOOSE-related results are averaged over iterations of \emph{present-feedback}. Base model is \texttt{GPT-3.5}.
}
\label{tab:future_past_feedback}
\end{table}
\begin{table}[]
\centering
\resizebox{1.0\columnwidth}{!}{
\begin{tabular}{l|ccc}
\toprule
                                     & Validness      & Novelty        & Helpfulness    \\ \midrule
MOOSE~(w/o \emph{present-feedback})   & 3.823          & 3.114          & 3.809          \\
\quad w/ 1 iteration of \emph{present-feedback}   & 3.918          & 3.199          & 3.900          \\
\quad w/ 2 iterations of \emph{present-feedback}   & 3.951          & 3.293          & 3.956          \\
\quad w/ 3 iterations of \emph{present-feedback}   & 3.969          & 3.270          & \textbf{3.962} \\
\quad w/ 4 iterations of \emph{present-feedback}   & \textbf{3.970} & \textbf{3.329} & 3.951          \\ 
\bottomrule
\end{tabular}}
\caption{Effect of \emph{present-feedback}~(evaluated by \texttt{GPT-4}). Base model is \texttt{GPT-3.5}.}
\label{tab:present}
\end{table}

In this subsection, we compare MOOSE-base with the baseline and examine the effect of each of the three feedback mechanisms to MOOSE-base.

We first introduce the number of generated hypotheses being evaluated in \S\ref{subsec:main_results} and \S\ref{sec:analysis}. 
For experiments evaluated with GPT-4, fifty backgrounds are selected for each method.
For MOOSE-related methods, for each background, on average around 6 inspirations are extracted, resulting in 4 different hypotheses.
Each hypothesis leads to another 4 more refined ones with present-feedback.
Therefore on average for each MOOSE-related method in GPT-4 evaluation tables, around 50*4*5=1000 hypotheses are evaluated.
For experiments evaluated with expert evaluation, in general, we randomly select one hypothesis for each background, resulting in 50 hypotheses evaluated for each line of the method in expert evaluation tables.

\begin{table}[]
\centering
\resizebox{1.0\columnwidth}{!}{
\begin{tabular}{l|ccc}
\toprule
                            & Validness     & Novelty       & Helpfulness   \\ \midrule
Baseline                    & 3.579          & 2.276          & 2.632          \\ \midrule
MOOSE-base                  & 3.500          & 2.855          & 3.026          \\
\quad w/ \emph{future-feedback}          & 3.645          & 3.105          & 3.303 \\
\quad w/ \emph{future-} and \emph{past-feedback} & \textbf{3.750} & \textbf{3.197} & \textbf{3.368}  
\\ \bottomrule
\end{tabular}}
\caption{Effect of MOOSE-base, \emph{future-feedback} and \emph{past-feedback}~(evaluated by \emph{experts}). MOOSE results are selected from the 5$^{th}$ iteration of \emph{present-feedback}. Base model is \texttt{GPT-3.5}.}
\label{tab:future_past_expert}
\end{table}
\begin{table}[]
\centering
\resizebox{1.0\columnwidth}{!}{
\begin{tabular}{l|ccc}
\toprule
                                         & Validness & Novelty & Helpfulness \\ \midrule
MOOSE-base~(w/o \emph{present-feedback})        & 3.342      & 2.382    & 2.500        \\
\quad w/ 2 iterations of \emph{present-feedback} & \textbf{3.539}      & 2.803    & 2.934        \\
\quad w/ 4 iterations of \emph{present-feedback} & 3.500      & \textbf{2.855}    & \textbf{3.026} 
\\ \midrule
MOOSE~(w/o \emph{present-feedback})             & 3.224      & 2.737    & 2.855        \\
\quad w/ 2 iterations of \emph{present-feedback} & 3.579      & \textbf{3.250}    & 3.342        \\
\quad w/ 4 iterations of \emph{present-feedback} & \textbf{3.750}      & 3.197    & \textbf{3.368}        \\ 
\bottomrule
\end{tabular}}
\caption{Effect of \emph{present-feedback}~(evaluated by \emph{experts}). Base model is \texttt{GPT-3.5}.}
\label{tab:present_expert}
\end{table}

Table~\ref{tab:future_past_feedback} shows GPT-4's evaluation targeting at comparing MOOSE-base and the baseline and shows the effect of future-feedback and past-feedback.
In this table, MOOSE-related results are averaged over iterations of present-feedback to not be influenced by present-feedback.
MOOSE-base largely outperforms the baseline in terms of both novelty and helpfulness, but slightly lower in terms of validness.
As illustrated in \S\ref{subsec:metrics}, since the purpose of the TOMATO task is to inspire and help human researchers, novelty and helpfulness metrics should be more important.
In practice, we find many hypotheses from baseline almost only rephrasing some sentences in the input corpus, adding little novelty content.
MOOSE-base with future-feedback comprehensively outperforms MOOSE-base in terms of all three metrics.
MOOSE-base with both future and past-feedback largely outperforms MOOSE-base with future-feedback in novelty and performs slightly lower in validness and helpfulness metrics.
One of the reasons is that the past-feedback may focus more on the novelty aspect because the novelty checker module provides more negative present-feedback than the reality checker module.

Table~\ref{tab:present} shows the effect of present-feedback with GPT-4 evaluation. 
In this table, the results are averaged over three experiments: MOOSE-base, MOOSE-base with future-feedback, and MOOSE-base with both future and past-feedback to focus on present-feedback.
It shows that as more iterations of present-feedback are conducted, validness and novelty steadily go up; helpfulness also steadily goes up but reaches the best performance with 3 iterations of present-feedback.

Table~\ref{tab:future_past_expert} shows expert evaluation results on the comparison between MOOSE-base and the baseline, and the effect of future-feedback and past-feedback.
MOOSE-related results are selected from the 5$^{th}$ iteration of present-feedback.
Similar to GPT-4 evaluation, 
MOOSE-base largely outperforms the baseline in terms of Novelty and Helpfulness;
MOOSE-base with future-feedback comprehensively outperforms MOOSE-base.
Different from GPT-4 evaluation, MOOSE-base with future and past-feedback also comprehensively outperforms MOOSE-base with future-feedback.
We think one of the reasons could be that GPT-4 might grade validness based on how frequently it has seen relevant texts, but not true understanding of the world.
Therefore a more novel hypothesis might tend to have a relatively lower score in validness and helpfulness under GPT-4 evaluation.

Table~\ref{tab:present_expert} shows the expert evaluation of present-feedback. 
MOOSE-base and MOOSE are both evaluated. 
Overall performance generally goes up with more iterations of present-feedback, but there might be an optimal number of iterations.

\section{Analysis}
\label{sec:analysis}

\subsection{Background and Inspirations}

Here we try to answer ``Is ChatGPT necessary for background and inspiration selection?''.

Table~\ref{tab:bkg_insp} shows various methods for background and inspiration selection.
In general, there might be a validness-novelty trade-off that if a method reaches a high novelty score, then it is usually hard for it to reach a high validness score.
It is surprising that a randomly selected background and randomly selected inspirations can lead to hypotheses with relatively comparable validness and novelty to ChatGPT-picked background and inspirations.
Empirically we hypothesize the reason is that randomly picked inspirations are mostly not related to the background, resulting in a high novelty~(but less validness and helpfulness).
In addition, BM25~\citep{robertson2009probabilistic} picked background and inspirations reach a much higher novelty score compared to ChatGPT-picked ones.
Empirically we do not find BM25 retrieved inspirations to be similar to the background, but they are usually with more concrete contents compared with random inspirations.
Not surprisingly, ChatGPT picked background and inspirations reach the highest helpfulness score among 
those without
any ground-truth annotations.
Lastly, ground-truth hypotheses reach the highest novelty and helpfulness.
\begin{table}[]
\centering
\resizebox{1.0\columnwidth}{!}{
\begin{tabular}{l|ccc}
\toprule
                                           & Validness      & Novelty & Helpfulness    \\ \midrule
Rand background                            & 3.954          & 2.483   & 3.489          \\
Rand background and rand inspirations      & 3.773          & 2.957   & 3.643          \\
Rand background and BM25 inspirations      & 3.585          & \textbf{3.364}   & 3.670          \\
GPT-3.5 picked background and inspirations & \textbf{3.812} & 2.818   & \textbf{3.733} \\ \midrule
Groundtruth background and inspirations         & \textbf{3.876} & 3.000   & 3.806          \\
Groundtruth hypotheses                          & 3.700          & \textbf{3.380}   & \textbf{3.880} \\ 
\bottomrule
\end{tabular}}
\caption{Analysis of retrieval's effect on generated hypotheses~(evaluated by \texttt{GPT-4}). No methods here utilize any feedback mechanisms. 
Base model is \texttt{GPT-3.5}.}
\label{tab:bkg_insp}
\end{table}

\subsection{More Ablation Studies}
\begin{table}[]
\centering
\resizebox{1.0\columnwidth}{!}{
\begin{tabular}{l|ccc}
\toprule
                             & Validness      & Novelty        & Helpfulness    \\ \midrule
MOOSE                        & 3.916          & 3.390          & 3.931          \\ 
\quad w/o \emph{future-feedback-2}        & 3.895          & 3.281          & 3.918          \\
\quad w/o \emph{future-feedback-1}        & 3.882          & 3.355          & 3.935          \\
\quad w/o access to related survey & 3.889          & \textbf{3.431} & 3.886          \\
\quad w/ randomized corpus         & \textbf{3.941} & 3.227          & \textbf{3.955} \\ 
\bottomrule
\end{tabular}}
\caption{More ablation study~(evaluated by \texttt{GPT-4}). 
Results are averaged over iterations of \emph{present-feedback}. Base model is \texttt{GPT-3.5}.}
\label{tab:more_ablation}
\end{table}

Table~\ref{tab:more_ablation} shows ablation studies on future-feedback, access to surveys, and the selection of corpus.

Firstly, for future-feedback, we separately test the effect of FF1 and FF2.
Without FF2, performance comprehensively drops; 
without FF1, performance drops on validness and novelty, with helpfulness remaining comparable. 
It seems that FF2 is more significant than FF1.
However, the fact that FF1 works on inspiration title finder and inspiration finer modules does not mean that it works on all modules.
Empirically we find that adding the reasons~(or prospects) for background and inspirations to the hypothesis proposer module will cause a more valid but much less novel generation of hypotheses.
The reason is that the hypothesis proposer module tends to simply follow the prospects, which do not have a global view of both background and all inspirations, but only focus on one background or one inspiration.
Instead, FF2~(the hypothesis suggestor module) has the global view and only provides soft initial suggestions on how to combine the background and inspirations together. 
With the hypotheses suggestor module, the hypotheses proposer module is prompted to further combine the initial suggestions and other inspirations to propose hypotheses.
To be fair, MOOSE-base, which is not equipped with the hypothesis suggestor module, has the same prompt to combine the inspirations together~(just without suggestions) to propose hypotheses.

\begin{table}[]
\centering
\resizebox{1.0\columnwidth}{!}{
\begin{tabular}{l|ccc}
\toprule
                            & Validness     & Novelty       & Helpfulness   \\ \midrule
Baseline                    & 	3.884          & 2.925          & 	3.856          \\ \midrule
MOOSE-base                  & \textbf{3.967}          & 3.392          & 3.939          \\
\quad w/ \emph{future-feedback}          & 3.926          & 3.694          & \textbf{3.966} \\
\quad w/ \emph{future-} and \emph{past-feedback} & 3.875 & \textbf{4.177} & 3.868  
\\ \bottomrule
\end{tabular}}
\caption{Effect of MOOSE-base, \emph{future-feedback} and \emph{past-feedback}~(evaluated by \texttt{GPT-4}). MOOSE-related results are averaged over iterations of \emph{present-feedback}. Base model is \texttt{Claude3-Opus}.}
\label{tab:future_past_gpt4_base_model_claude3}
\end{table}
Secondly, we cut the access of novelty detector to related surveys to check the effect of related surveys. 
As a result, novelty largely goes up~(0.04), and validness goes down to around 0.26.
Empirically one of the main reasons is that BM25 hardly retrieves enough similar survey chunks, so that access to the survey leads novelty detector to tend to reply the hypotheses are novel since it is not mentioned in the related survey.
Without present-feedback, MOOSE and MOOSE w/o access to survey perform quite comparably.

Lastly, the raw corpus in the dataset is from two sources: passages that contain the ground truth backgrounds and passages that contain the ground truth inspirations. 
In all of the previous experiments, backgrounds are extracted from the background passages, and inspirations are extracted from the inspirations passages.
To see whether the passages are only restricted to their designed role, in MOOSE w/ randomized corpus experiment, we use inspiration corpus for background extraction and use both inspiration and background corpus for inspiration extraction.
As a result, validness goes up by about 0.025, while novelty goes down by about 0.16. 
We think one of the reasons is that, in this setting, after selecting a background from an inspiration passage, MOOSE tends to retrieve the same inspiration passage to find inspirations, which leads to less novel results.

\begin{table}[]
\centering
\resizebox{1.0\columnwidth}{!}{
\begin{tabular}{l|ccc}
\toprule
                                     & Validness      & Novelty        & Helpfulness    \\ \midrule
MOOSE~(w/o \emph{present-feedback})   & 3.793          & 3.683          & 3.870          \\
\quad w/ 1 iteration of \emph{present-feedback}   & 3.896          & 3.804          & 3.937          \\
\quad w/ 2 iterations of \emph{present-feedback}   & 3.961          & 3.730          & 3.939          \\
\quad w/ 3 iterations of \emph{present-feedback}   & \textbf{3.983}          & \textbf{3.809}          & \textbf{3.946} \\
\quad w/ 4 iterations of \emph{present-feedback}   & 3.980            &         3.757      &      3.930          \\ 
\bottomrule
\end{tabular}}
\caption{Effect of \emph{present-feedback}~(evaluated by \texttt{GPT-4}). Base model is \texttt{Claude3-Opus}.}
\label{tab:present_claude3}
\end{table}

\subsection{Effect of Base Model Selection}
\label{sec:base_model_selection}

In all previous experiments, we adopt \texttt{GPT-3.5} as the base model. 
In this section, we investigate the effect of base model selection by using \texttt{Claude3-Opus} as the base model for each module in MOOSE.

With \texttt{Claude3-Opus} as the base model, we again analyze the effect of MOOSE-base, past-feedback, and future-feedback in Table~\ref{tab:future_past_gpt4_base_model_claude3}; and analyze the effect of present-feedback in Table~\ref{tab:present_claude3}.
The experiment settings of Table~\ref{tab:future_past_gpt4_base_model_claude3} and Table~\ref{tab:present_claude3} are exactly the same as in Table~\ref{tab:future_past_feedback} and Table~\ref{tab:present} correspondingly, but only differ in the base model selection.

In general, there are two conclusions.
Firstly, MOOSE's components stay effective regardless of different base model selection. 
It shows the robustness of the MOOSE framework in terms of different base model. 
Secondly, the absolute evaluation scores on all three metrics largely improved with \texttt{Claude3-Opus} compared to \texttt{GPT-3.5}, indicating the even larger potential of the MOOSE framework when more powerful LLMs are available.


\subsection{Qualitative Analysis}
\label{sec:qualitative_analysis}

The following box shows one generated counter-intuitive hypothesis~(expert evaluation appended).

\mybox{gray!10}{\textit{In collectivist cultures, individuals engage in more conspicuous consumption behaviors compared to individualistic cultures.}~(Validness:~3.3; Novelty:~4.0; Helpfulness:~4.0)}

Here is the assessment from one of the experts:

\emph{The main reason I give a high mark for both three dimensions of this hypothesis is because:}

\emph{(1) For validness, this hypothesis is based on existing cultural theories and empirical evidence that suggests cultural values significantly impact consumer behavior. It aligns with established concepts like collectivism and individualism that have been widely studied in cross-cultural psychology.}

\emph{(2) For novelty, this hypothesis is counter-intuitive to some extent. Prior research has shown that collectivist cultures often prioritize group harmony, cooperation, and social cohesion over individual desires.
This emphasis on collective well-being might suggest a reduced inclination toward overt displays of personal wealth or status through conspicuous consumption. However, this hypothesis suggests the opposite 
that
collectivist culture's members
engage in more conspicuous consumption, which is more commonly linked to individualistic societies in popular perceptions. This challenges the notion that members of collectivist cultures avoid conspicuous consumption behaviors.}

\emph{(3) For helpfulness, if this hypothesis is confirmed, it could have significant practical implications. Understanding the impact of cultural values on conspicuous consumption can assist businesses and marketers in crafting more effective cross-cultural marketing strategies. It could also aid policymakers in addressing societal issues related to consumerism.} 

In addition to the analysis of this counter-intuitive example, we also provide qualitative analysis on the difference between hypotheses generated from the baseline, MOOSE-base, MOOSE-base w/ future-feedback, and MOOSE-base w/ future and past-feedback in \S\ref{appen:qualitative_analysis}.
More qualitative analysis on highly scored generated hypotheses can be found in \S\ref{appen:old_two_qualitative_analysis}. Additionally, \S\ref{appen:factors_for_good_hypotheses} illustrates factors for good hypotheses in social science~(particularly in Business).
\S\ref{appen:an_example_hypothesis_formulation} shows how MOOSE formulates a hypothesis by giving the generation of each of the modules in MOOSE.


\section{Conclusion}
In this paper, we propose a new task, automated open-domain hypothetical induction~(TOMATO), which is the first task in NLP to focus on social science
research hypotheses discovery.
Along with the task, we construct a dataset consisting of 50 recent social science
papers published in top academic journals.
We also developed a multi-module framework MOOSE for the TOMATO task, which contains a base framework and three novel feedback mechanisms.
Experiments indicate that MOOSE-base outperforms an LLM-based baseline, and the three feedback mechanisms can progressively further improve over MOOSE-base. 
Surprisingly, evaluated by PhD students, MOOSE is able to produce many novel~(``not existing in the literature'') and valid~(``reflecting reality'') research hypotheses.
To the best of our knowledge, this is the first work showing that LLMs can be leveraged to generate novel and valid scientific hypotheses, 
indicating the potential of LLMs to serve as a ``copilot'' for scientists.

\section*{Limitations}
From the first look, it might seem that the proposed dataset consists of only 50 recent papers.
However, they are all manually collected by experts~(PhD students), and are annotated with lots of details~(e.g., identifying background and inspirations, finding relevant raw web passages for background and inspirations, reasoning process, complexity level).
In addition, each paper has been published in a top social science journal, representing the pinnacle of human intelligence. This means it would be incredibly exciting if LLMs could propose a hypothesis from even a single one of these recent papers.

It might also seem that it is not clear whether the design of the framework can apply to other disciplines. 
However, to the best of our knowledge, this is the first paper using LLMs that can propose novel scientific hypotheses that are new to humanity. 
We choose social science as the breakthrough point since the main data format of social science is language.
Table~\ref{tab:dataset_subject} shows that the dataset covers 7 different disciplines (e.g., Psychology, Management, Marketing).
It would be nearly impossible for the first few works to develop a general method to propose novel hypotheses for all disciplines.

This paper concentrates on an automated task setting in which a system is designed to formulate scientific hypotheses independently, without requiring human intervention.
In some scenarios, scientists may prefer to use their own background and inspirations as input for controllable hypotheses generation. 
It might seem that the automated setting and the controllable setting are in conflict.
However, we contend that the automated setting make a step further than the controllable setting, since a system developed for an automated setting would inherently support controllable generation by simply substituting the automatically searched inputs~(e.g., background and inspirations) with those that are manually crafted.

\textbf{Societal Impact}: Expert evaluation shows that MOOSE, an LLM-based system, might already be able to serve as a copilot for researchers across various social science disciplines.
Particularly, as depicted in Figure~\ref{fig:intro_figure}, it can assist researchers in the hypothesis formation process, which is the first step for scientific discovery~\citep{wang2023scientific}.
This capability signifies a step towards enhancing the efficiency of scientific exploration by accelerating the formation and development of innovative and credible research hypotheses, thereby boosting researchers' productivity. 
To maximize its impact and ensure equitable access, it is imperative to advocate for the open-sourcing of such systems, thereby democratizing access for the global scientific community.


\section*{Acknowledgement}
This research/project is supported by the Ministry of Education, Singapore under its MOE Academic Research Fund Tier 2 (STEM RIE2025 Award MOE-T2EP20123-0005).
We thank Qingyun Wang, Jinjie Ni, Xulang Zhang, and Qika Lin for their insightful comments on the first finished draft of this work.

\bibliography{anthology,custom}

\begin{thebibliography}{29}
\expandafter\ifx\csname natexlab\endcsname\relax\def\natexlab#1{#1}\fi

\bibitem[{Anthropic(2024)}]{anthropic2024claude}
AI~Anthropic. 2024.
\newblock The claude 3 model family: Opus, sonnet, haiku.
\newblock \emph{Claude-3 Model Card}.

\bibitem[{Banerjee and Lavie(2005)}]{banerjee-lavie-2005-meteor}
Satanjeev Banerjee and Alon Lavie. 2005.
\newblock \href {https://aclanthology.org/W05-0909} {{METEOR}: An automatic metric for {MT} evaluation with improved correlation with human judgments}.
\newblock In \emph{Proceedings of the {ACL} Workshop on Intrinsic and Extrinsic Evaluation Measures for Machine Translation and/or Summarization}, pages 65--72, Ann Arbor, Michigan. Association for Computational Linguistics.

\bibitem[{Boiko et~al.(2023)Boiko, MacKnight, and Gomes}]{DBLP:journals/corr/abs-2304-05332}
Daniil~A. Boiko, Robert MacKnight, and Gabe Gomes. 2023.
\newblock \href {https://doi.org/10.48550/arXiv.2304.05332} {Emergent autonomous scientific research capabilities of large language models}.
\newblock \emph{CoRR}, abs/2304.05332.

\bibitem[{Bosselut et~al.(2019)Bosselut, Rashkin, Sap, Malaviya, Celikyilmaz, and Choi}]{bosselut-etal-2019-comet}
Antoine Bosselut, Hannah Rashkin, Maarten Sap, Chaitanya Malaviya, Asli Celikyilmaz, and Yejin Choi. 2019.
\newblock \href {https://doi.org/10.18653/v1/P19-1470} {{COMET}: Commonsense transformers for automatic knowledge graph construction}.
\newblock In \emph{Proceedings of the 57th Annual Meeting of the Association for Computational Linguistics}, pages 4762--4779, Florence, Italy. Association for Computational Linguistics.

\bibitem[{Bran et~al.(2023)Bran, Cox, White, and Schwaller}]{bran2023chemcrow}
Andres~M Bran, Sam Cox, Andrew~D White, and Philippe Schwaller. 2023.
\newblock Chemcrow: Augmenting large-language models with chemistry tools.
\newblock \emph{arXiv preprint arXiv:2304.05376}.

\bibitem[{Das et~al.(2021)Das, Zaheer, Thai, Godbole, Perez, Lee, Tan, Polymenakos, and McCallum}]{das-etal-2021-case}
Rajarshi Das, Manzil Zaheer, Dung Thai, Ameya Godbole, Ethan Perez, Jay~Yoon Lee, Lizhen Tan, Lazaros Polymenakos, and Andrew McCallum. 2021.
\newblock \href {https://doi.org/10.18653/v1/2021.emnlp-main.755} {Case-based reasoning for natural language queries over knowledge bases}.
\newblock In \emph{Proceedings of the 2021 Conference on Empirical Methods in Natural Language Processing}, pages 9594--9611, Online and Punta Cana, Dominican Republic. Association for Computational Linguistics.

\bibitem[{Gao et~al.(2023)Gao, Rong, Tian, and Yao}]{gao2023improving}
Jia Gao, Ying Rong, Xin Tian, and Yuliang Yao. 2023.
\newblock Improving convenience or saving face? an empirical analysis of the use of facial recognition payment technology in retail.
\newblock \emph{Information Systems Research}.

\bibitem[{Goel et~al.(2017)Goel, Navarrete, Noveck, and Prado}]{goel2017reasoning}
Vinod Goel, Gorka Navarrete, Ira~A Noveck, and J{\'e}r{\^o}me Prado. 2017.
\newblock The reasoning brain: The interplay between cognitive neuroscience and theories of reasoning.

\bibitem[{Lin(2004)}]{lin2004rouge}
Chin-Yew Lin. 2004.
\newblock Rouge: A package for automatic evaluation of summaries.
\newblock In \emph{Text summarization branches out}, pages 74--81.

\bibitem[{Liu et~al.(2016)Liu, Lowe, Serban, Noseworthy, Charlin, and Pineau}]{liu-etal-2016-evaluate}
Chia-Wei Liu, Ryan Lowe, Iulian Serban, Mike Noseworthy, Laurent Charlin, and Joelle Pineau. 2016.
\newblock \href {https://doi.org/10.18653/v1/D16-1230} {How {NOT} to evaluate your dialogue system: An empirical study of unsupervised evaluation metrics for dialogue response generation}.
\newblock In \emph{Proceedings of the 2016 Conference on Empirical Methods in Natural Language Processing}, pages 2122--2132, Austin, Texas. Association for Computational Linguistics.

\bibitem[{Madaan et~al.(2023)Madaan, Tandon, Gupta, Hallinan, Gao, Wiegreffe, Alon, Dziri, Prabhumoye, Yang, Welleck, Majumder, Gupta, Yazdanbakhsh, and Clark}]{DBLP:journals/corr/abs-2303-17651}
Aman Madaan, Niket Tandon, Prakhar Gupta, Skyler Hallinan, Luyu Gao, Sarah Wiegreffe, Uri Alon, Nouha Dziri, Shrimai Prabhumoye, Yiming Yang, Sean Welleck, Bodhisattwa~Prasad Majumder, Shashank Gupta, Amir Yazdanbakhsh, and Peter Clark. 2023.
\newblock \href {https://doi.org/10.48550/arXiv.2303.17651} {Self-refine: Iterative refinement with self-feedback}.
\newblock \emph{CoRR}, abs/2303.17651.

\bibitem[{Norton(2003)}]{norton2003little}
John~D Norton. 2003.
\newblock A little survey of induction.

\bibitem[{OpenAI(2023)}]{DBLP:journals/corr/abs-2303-08774}
OpenAI. 2023.
\newblock \href {https://doi.org/10.48550/arXiv.2303.08774} {{GPT-4} technical report}.
\newblock \emph{CoRR}, abs/2303.08774.

\bibitem[{Ouyang et~al.(2022)Ouyang, Wu, Jiang, Almeida, Wainwright, Mishkin, Zhang, Agarwal, Slama, Ray, Schulman, Hilton, Kelton, Miller, Simens, Askell, Welinder, Christiano, Leike, and Lowe}]{DBLP:conf/nips/Ouyang0JAWMZASR22}
Long Ouyang, Jeffrey Wu, Xu~Jiang, Diogo Almeida, Carroll~L. Wainwright, Pamela Mishkin, Chong Zhang, Sandhini Agarwal, Katarina Slama, Alex Ray, John Schulman, Jacob Hilton, Fraser Kelton, Luke Miller, Maddie Simens, Amanda Askell, Peter Welinder, Paul~F. Christiano, Jan Leike, and Ryan Lowe. 2022.
\newblock \href {http://papers.nips.cc/paper\_files/paper/2022/hash/b1efde53be364a73914f58805a001731-Abstract-Conference.html} {Training language models to follow instructions with human feedback}.
\newblock In \emph{NeurIPS}.

\bibitem[{Pan et~al.(2011)Pan, Mulkar{-}Mehta, and Hobbs}]{DBLP:journals/coling/PanMH11}
Feng Pan, Rutu Mulkar{-}Mehta, and Jerry~R. Hobbs. 2011.
\newblock \href {https://doi.org/10.1162/COLI\_a\_00075} {Annotating and learning event durations in text}.
\newblock \emph{Comput. Linguistics}, 37(4):727--752.

\bibitem[{Papineni et~al.(2002)Papineni, Roukos, Ward, and Zhu}]{papineni2002bleu}
Kishore Papineni, Salim Roukos, Todd Ward, and Wei-Jing Zhu. 2002.
\newblock Bleu: a method for automatic evaluation of machine translation.
\newblock In \emph{Proceedings of the 40th annual meeting of the Association for Computational Linguistics}, pages 311--318.

\bibitem[{Peng et~al.(2023)Peng, Galley, He, Cheng, Xie, Hu, Huang, Liden, Yu, Chen, and Gao}]{DBLP:journals/corr/abs-2302-12813}
Baolin Peng, Michel Galley, Pengcheng He, Hao Cheng, Yujia Xie, Yu~Hu, Qiuyuan Huang, Lars Liden, Zhou Yu, Weizhu Chen, and Jianfeng Gao. 2023.
\newblock \href {https://doi.org/10.48550/arXiv.2302.12813} {Check your facts and try again: Improving large language models with external knowledge and automated feedback}.
\newblock \emph{CoRR}, abs/2302.12813.

\bibitem[{Press et~al.(2022)Press, Zhang, Min, Schmidt, Smith, and Lewis}]{DBLP:journals/corr/abs-2210-03350}
Ofir Press, Muru Zhang, Sewon Min, Ludwig Schmidt, Noah~A. Smith, and Mike Lewis. 2022.
\newblock \href {https://doi.org/10.48550/arXiv.2210.03350} {Measuring and narrowing the compositionality gap in language models}.
\newblock \emph{CoRR}, abs/2210.03350.

\bibitem[{Qi et~al.(2023)Qi, Zhang, Li, Tian, Zeng, Chen, and Zhou}]{DBLP:journals/corr/abs-2311-05965}
Biqing Qi, Kaiyan Zhang, Haoxiang Li, Kai Tian, Sihang Zeng, Zhang{-}Ren Chen, and Bowen Zhou. 2023.
\newblock \href {https://doi.org/10.48550/ARXIV.2311.05965} {Large language models are zero shot hypothesis proposers}.
\newblock \emph{CoRR}, abs/2311.05965.

\bibitem[{Robertson et~al.(2009)Robertson, Zaragoza et~al.}]{robertson2009probabilistic}
Stephen Robertson, Hugo Zaragoza, et~al. 2009.
\newblock The probabilistic relevance framework: Bm25 and beyond.
\newblock \emph{Foundations and Trends{\textregistered} in Information Retrieval}, 3(4):333--389.

\bibitem[{Shinn et~al.(2023)Shinn, Labash, and Gopinath}]{DBLP:journals/corr/abs-2303-11366}
Noah Shinn, Beck Labash, and Ashwin Gopinath. 2023.
\newblock \href {https://doi.org/10.48550/arXiv.2303.11366} {Reflexion: an autonomous agent with dynamic memory and self-reflection}.
\newblock \emph{CoRR}, abs/2303.11366.

\bibitem[{Wang et~al.(2023{\natexlab{a}})Wang, Fu, Du, Gao, Huang, Liu, Chandak, Liu, Van~Katwyk, Deac et~al.}]{wang2023scientific}
Hanchen Wang, Tianfan Fu, Yuanqi Du, Wenhao Gao, Kexin Huang, Ziming Liu, Payal Chandak, Shengchao Liu, Peter Van~Katwyk, Andreea Deac, et~al. 2023{\natexlab{a}}.
\newblock Scientific discovery in the age of artificial intelligence.
\newblock \emph{Nature}, 620(7972):47--60.

\bibitem[{Wang et~al.(2023{\natexlab{b}})Wang, Downey, Ji, and Hope}]{DBLP:journals/corr/abs-2305-14259}
Qingyun Wang, Doug Downey, Heng Ji, and Tom Hope. 2023{\natexlab{b}}.
\newblock \href {https://doi.org/10.48550/arXiv.2305.14259} {Learning to generate novel scientific directions with contextualized literature-based discovery}.
\newblock \emph{CoRR}, abs/2305.14259.

\bibitem[{Yang et~al.(2022)Yang, Tian, Peng, and Klein}]{DBLP:conf/emnlp/YangTPK22}
Kevin Yang, Yuandong Tian, Nanyun Peng, and Dan Klein. 2022.
\newblock \href {https://doi.org/10.18653/v1/2022.emnlp-main.296} {Re3: Generating longer stories with recursive reprompting and revision}.
\newblock In \emph{Proceedings of the 2022 Conference on Empirical Methods in Natural Language Processing, {EMNLP} 2022, Abu Dhabi, United Arab Emirates, December 7-11, 2022}, pages 4393--4479. Association for Computational Linguistics.

\bibitem[{Yang et~al.(2024)Yang, Dong, Du, Cheng, Cambria, Liu, Gao, and Wei}]{yang-etal-2024-language}
Zonglin Yang, Li~Dong, Xinya Du, Hao Cheng, Erik Cambria, Xiaodong Liu, Jianfeng Gao, and Furu Wei. 2024.
\newblock \href {https://aclanthology.org/2024.eacl-long.13} {Language models as inductive reasoners}.
\newblock In \emph{Proceedings of the 18th Conference of the European Chapter of the Association for Computational Linguistics (Volume 1: Long Papers)}, pages 209--225, St. Julian{'}s, Malta. Association for Computational Linguistics.

\bibitem[{Yang et~al.(2023{\natexlab{a}})Yang, Du, Cambria, and Cardie}]{yang-etal-2023-end}
Zonglin Yang, Xinya Du, Erik Cambria, and Claire Cardie. 2023{\natexlab{a}}.
\newblock \href {https://aclanthology.org/2023.eacl-main.255} {End-to-end case-based reasoning for commonsense knowledge base completion}.
\newblock In \emph{Proceedings of the 17th Conference of the European Chapter of the Association for Computational Linguistics}, pages 3509--3522, Dubrovnik, Croatia. Association for Computational Linguistics.

\bibitem[{Yang et~al.(2023{\natexlab{b}})Yang, Du, Mao, Ni, and Cambria}]{DBLP:journals/corr/abs-2303-12023}
Zonglin Yang, Xinya Du, Rui Mao, Jinjie Ni, and Erik Cambria. 2023{\natexlab{b}}.
\newblock \href {https://doi.org/10.48550/arXiv.2303.12023} {Logical reasoning over natural language as knowledge representation: {A} survey}.
\newblock \emph{CoRR}, abs/2303.12023.

\bibitem[{Yang et~al.(2020)Yang, Du, Rush, and Cardie}]{yang-etal-2020-improving}
Zonglin Yang, Xinya Du, Alexander Rush, and Claire Cardie. 2020.
\newblock \href {https://doi.org/10.18653/v1/2020.findings-emnlp.302} {Improving event duration prediction via time-aware pre-training}.
\newblock In \emph{Findings of the Association for Computational Linguistics: EMNLP 2020}, pages 3370--3378, Online. Association for Computational Linguistics.

\bibitem[{Zhong et~al.(2023)Zhong, Zhang, Li, Ahn, Klein, and Steinhardt}]{DBLP:journals/corr/abs-2302-14233}
Ruiqi Zhong, Peter Zhang, Steve Li, Jinwoo Ahn, Dan Klein, and Jacob Steinhardt. 2023.
\newblock \href {https://doi.org/10.48550/arXiv.2302.14233} {Goal driven discovery of distributional differences via language descriptions}.
\newblock \emph{CoRR}, abs/2302.14233.

\end{thebibliography}
\bibliographystyle{acl_natbib}

\FloatBarrier

\newpage
\appendix
\onecolumn

\section{Appendix}
\label{sec:appendix}

\subsection{Hyper-parameters}
\label{appen:hyperparameters}
Experiments in \S\ref{sec:base_model_selection} adopts \texttt{Claude3-Opus}, all other experiments are conducted with \texttt{gpt-3.5-turbo}.

Both \texttt{Claude3-Opus} and \texttt{gpt-3.5-turbo} use 0.9 temperature and 0.9 top\_p. 

The hyperparameters for \texttt{GPT-4} evaluation are 0.0 temperature to ensure the evaluation scores are stable, and 0.9 top\_p.



\subsection{More Related Works on Reasoning and Scientific Discovery}
\label{appen:more_related_works}

This paper is a successive work in inductive reasoning and is different from commonsense reasoning~\citep{bosselut-etal-2019-comet,yang-etal-2020-improving} in that the novel social science hypotheses do not belong to commonsense. 

Case-based reasoning~\citep{das-etal-2021-case,yang-etal-2023-end} also falls in the domain of inductive reasoning, but case-based reasoning is more about high-level guidance on methodology design~(case retrieve, reuse, revise, and retain), which is not involved in this paper.

\citet{DBLP:journals/corr/abs-2311-05965} work on zero-shot hypothesis proposing, which is a concurrent work to our paper. 
They don't focus on social science and business disciplines, and mostly adopt a single LLM as method~(prompting, finetuning).

\FloatBarrier
\subsection{Dataset Complexity Distribution}
\label{appen:dataset_conplexity}
\begin{table}[]
\centering
\resizebox{0.5\columnwidth}{!}{
\begin{tabular}{l|cc}
\toprule
       & Reasoning Complexity                     & Association Complexity            \\ \midrule
Easy   & 24                                       & 12                                \\
Medium & 17                                       & 25                                \\
Hard   & 9                                        & 13                                 \\
\bottomrule
\end{tabular}
}
\caption{Statistics of the complexity of the dataset.}
\label{tab:dataset_complexity}
\end{table}
Table~\ref{tab:dataset_complexity} illustrates the complexity distribution of the proposed dataset from both reasoning and association perspectives.
``Easy'' in the table means it is relatively easy compared to other publications in the dataset, but does not mean it is actually easy to induce the hypotheses.

\subsection{Why the Tomato Dataset Shouldn’t Be Collected by Automatic Methods}
\label{appen:why_cant_automatic_collect_dataset}

Firstly, there are many hypotheses in a social science publication, which might need an expert to identify which hypothesis is suitable for this task~(e.g., whether it is a main hypothesis, whether the background and inspirations are properly introduced).

Secondly, the background and inspirations scatter in a publication. 
It needs a deep domain understanding of the hypothesis, related background, and inspirations to select the background and inspirations out to form a complete reasoning chain to conclude the hypothesis.

Thirdly, it needs enough domain knowledge to find semantically similar texts~(similar to the groundtruth selected background and inspirations) from the web, where the texts should contain enough details to help elicit the hypothesis.

\subsection{Why Not Using Other Evaluation Metrics}
\label{appen:why_not_other_eval_metrics}
Other relevant aspects from related literature include relevance~\citep{DBLP:journals/corr/abs-2305-14259} and significance~\citep{DBLP:journals/corr/abs-2302-14233}.

We do not adopt relevance because our task setting is the automated and open domain, without a manually given background; neither for significance because
social science is different from engineering subjects --- 
(1) every hypothesis is to reflect the reality of the world, and as long as it reflects the world, it is significant. 
Therefore it is hard to tell which one is more significant even by experts;
(2) the evaluation standard of significance varies from time to time. 
For example, in the 60s, conducting research on how to improve the assembly line's efficiency as much as possible was seen as very significant.
However, in recent decades, how to alleviate the psychological depression of assembly line workers is seen as more significant.

We do not adopt BLEU~\cite{papineni2002bleu}, ROUGE~\cite{lin2004rouge}, or METEOR~\citep{banerjee-lavie-2005-meteor} as evaluation metric to compare the proposed hypothesis and the ground truth hypothesis since (1) proposing novel research hypotheses is an open problem, and (2) TOMATO has an automated open domain setting, which means the automatically selected background and inspirations are hardly the same as a few given ground truth ones~(if background and inspirations are not the same, then it is meaningless to compare the hypothesis).
\citet{liu-etal-2016-evaluate} have conducted a comprehensive analysis that they also reached a similar conclusion that BLEU, METEOR, or ROUGE is not suitable for an open-ended task (such as a dialogue system).


\subsection{Hypotheses Selection for Expert Evaluation}
\label{appen:how_to_select_400_hypotheses}

In total, we randomly selected 400 hypotheses to be evaluated by experts.
Specifically, for each background passage in the dataset~(out of 50), we use 4 methods~(which are to be compared) to collect in total 8 hypotheses.

The 8 hypotheses are from (1) the baseline; (2) the MOOSE-base framework; (3) MOOSE-base + future-feedback; (4) MOOSE-base + future-feedback + past-feedback.
For (2) and (4), we collect three hypotheses, which are (a) without present-feedback; (b) after 2 iterations of present-feedback; and (c) after 4 iterations of present-feedback.
For (1) and (3), we only collect one hypothesis, which is without present-feedback.

With these collections, we can evaluate the effect of both the MOOSE-base framework and the three feedback methods, leading to results in Table~\ref{tab:future_past_expert} and Table~\ref{tab:present_expert}.

Out of the three experts, one expert evaluates the full 400 hypotheses, and the other two each evaluate 104 hypotheses~(the first and second 104 hypotheses out of 400). 
The reason we choose the number ``104'' is that 
(1) social science PhD students are quite busy and two of them can only have time to evaluate around 100 hypotheses;
(2) the number should be dividable by 8~(since every 8 hypotheses form a group for comparison).

The results of the expert evaluation are averaged over the three experts.
Specifically, expert evaluation essentially compares the 8 hypotheses within a group. 
The 400, 104, and 104 hypotheses evaluation scores can be written as arrays of [50, 8], [13, 8], and [13, 8]. We concatenate them to [76, 8], and average them across the first dimension.

The payment for expert evaluation is \$1 per hypothesis.

\subsection{Expert Qualification and Expert Agreement}
\label{appen:expert_qualification_and_agreement}
\begin{table}[]
\centering
\resizebox{0.5\columnwidth}{!}{
\begin{tabular}{l|ccc}
\toprule
            & Validness & Novelty & Helpfulness \\ \midrule
Hard Consistency & 0.298     & 0.337   & 0.361   \\
Soft Consistency & 0.755     & 0.793   & 0.791    \\ 
\bottomrule
\end{tabular}}
\caption{Hard and soft consistency scores between evaluation from different experts in terms of Validness, Novelty, and Helpfulness metrics.}
\label{tab:consistency_between_expert}
\end{table}
The constructed dataset covers many subjects, but every collected publication is somewhat related to Marketing, which is a big topic in Business research. 
It is common in social science to conduct research that connects with other social science domains.
The experts for expert evaluation are three PhD students majoring in Marketing.
%
Therefore the experts are qualified enough to provide assessment for machine-generated hypotheses in the domain.

The consistency scores between experts are shown in Table~\ref{tab:consistency_between_expert}.
The soft consistency and hard consistency are defined in \S\ref{appen:consistency_expert_gpt4}.
All soft consistency scores are above 0.75 means, and the average difference between experts in terms of each metric is less than 1~(out of a 5-point scale), exhibiting high expert evaluation agreement.

\begin{table}[]
\centering
\resizebox{0.5\columnwidth}{!}{
\begin{tabular}{l|ccc}
\toprule
            & Validness & Novelty & Helpfulness \\ \midrule
Hard Consistency & 0.485     & 0.392   & 0.321   \\
Soft Consistency & 0.850     & 0.823   & 0.773    \\ 
\bottomrule
\end{tabular}}
\caption{Hard and soft consistency scores between expert evaluation and GPT-4 evaluation in terms of Validness, Novelty, and Helpfulness metrics.}
\label{tab:consistency}
\end{table}
\subsection{Consistency Between Expert Evaluation and GPT-4 Evaluation}
\label{appen:consistency_expert_gpt4}
To check the consistency between expert evaluation and GPT-4 evaluation, we use the expert evaluation results and find the corresponding GPT-4 evaluation results.
In total, there are 400 hypotheses evaluated by experts, so the sample we use to calculate the consistency score is 400.

\begin{table}[]
\centering
\resizebox{0.6\columnwidth}{!}{
\begin{tabular}{l|l}
\toprule
\multicolumn{2}{c}{Aspect 1: Validness}                                                                                                                                \\ \midrule
5 points & \parbox{0.5\linewidth}{\centering The hypothesis completely reflects the reality.}                                                                                                              \\ \midrule
4 points & \parbox{0.5\linewidth}{\centering The hypothesis almost completely reflects the reality, but has only one or two minor conflictions that can be easily modified.}                               \\ \midrule
3 points & \parbox{0.5\linewidth}{\centering The hypothesis has at least one moderate conflict or several minor conflicts.}                                                                                \\ \midrule
2 points & \parbox{0.5\linewidth}{\centering The hypothesis has at least one major confliction with the reality or only establishes in very rare circumstances that are not mentioned in this hypothesis.} \\ \midrule
1 point  & \parbox{0.5\linewidth}{\centering The hypothesis completely violates the reality.}        \\ \bottomrule                                                                                                     
\end{tabular}}
\caption{Evaluation standard for \emph{Validness}.}
\label{tab:eval_validness}
\end{table}

\begin{table}[]
\centering
\resizebox{0.6\columnwidth}{!}{
\begin{tabular}{l|l}
\toprule
\multicolumn{2}{c}{Aspect 1: Novelty}                                                                                                                                \\ \midrule
5 points & \parbox{0.5\linewidth}{\centering The hypothesis is completely novel and has not been proposed by any existing literature.}                                                                                                              \\ \midrule
4 points & \parbox{0.5\linewidth}{\centering The main argument or several sub-arguments of the hypothesis are novel.}                               \\ \midrule
3 points & \parbox{0.5\linewidth}{\centering The main argument is not novel, only one or two sub-arguments appear to be novel.}                                                                                \\ \midrule
2 points & \parbox{0.5\linewidth}{\centering The full hypothesis is not novel, but the way it combines the topics can be inspiring for human researchers.} \\ \midrule
1 point  & \parbox{0.5\linewidth}{\centering The hypothesis is not novel at all and not inspiring for human researchers.}        \\ \bottomrule                                                                                                     
\end{tabular}}
\caption{Evaluation standard for \emph{Novelty}.}
\label{tab:eval_novelty}
\end{table}

\begin{table}[]
\centering
\resizebox{0.6\columnwidth}{!}{
\begin{tabular}{l|l}
\toprule
\multicolumn{2}{c}{Aspect 1: Helpfulness}                                                                                                                                \\ \midrule
5 points & \parbox{0.5\linewidth}{\centering The hypothesis is novel, valid, clear, and specific enough that it is itself a mature research hypothesis, and human researchers can directly adopt it for publication with no modifications needed.}                                                                                                              \\ \midrule
4 points & \parbox{0.5\linewidth}{\centering The hypothesis is novel enough and can be directly adopted by human researchers for publication after minor modifications.}                               \\ \midrule
3 points & \parbox{0.5\linewidth}{\centering The hypothesis should be largely modified or reconstructed by human researchers to adopt it.}                                                                                \\ \midrule
2 points & \parbox{0.5\linewidth}{\centering Modifying this hypothesis might not deserve the efforts, but a small part of this hypothesis is inspiring for human researchers to develop a new hypothesis.} \\ \midrule
1 point  & \parbox{0.5\linewidth}{\centering The hypothesis is not helpful and not inspiring at all.}        \\ \bottomrule                                                                                                     
\end{tabular}}
\caption{Evaluation standard for \emph{Helpfulness}.}
\label{tab:eval_helpfulness}
\end{table}
Specifically,
similar to ~\citet{DBLP:journals/coling/PanMH11},
for soft consistency, if the absolute difference between expert evaluation and GPT-4 evaluation~(both are on a 5-point scale) is 0/1/2/3/4, then we assign a consistency score of 1.00/0.75/0.50/0.25/0.00; 
for hard consistency, if only the difference is 0, can the consistency score be 1.00, otherwise consistency score is 0.00.
The hard and soft consistency scores shown in Table~\ref{tab:consistency} are averaged for each metric.

The consistency scores are surprisingly high. 
All soft consistency scores are above 0.75 means, and the average difference between expert and GPT-4 evaluation in terms of each metric is less than 1~(out of a 5-point scale).
The results indicate that GPT-4 might be able to provide a relatively reliable evaluation for machine-generated hypotheses.

\subsection{Evaluation Aspects Description}
\label{appen:eval_metrics_description}

The evaluation standard for \emph{Validness}, \emph{Novelty}, and \emph{Helpfulness} is correspondingly displayed in Table~\ref{tab:eval_validness}, Table~\ref{tab:eval_novelty}, and Table~\ref{tab:eval_helpfulness}.

\subsection{More Details About Past-Feedback Design}
\label{appen:detail_past_feedback}
In practice, we find that ChatGPT is not capable enough to generate past-feedback with enough good quality for the Inspiration Feedback module. 
Instead, it tends to provide feedback as ``the previous inspiration titles are not very relevant to the hypotheses or the background''.
As a result, the ChatGPT Inspiration Title Finder module tends to select inspiration titles that are very related to the background, resulting in a less novel hypotheses generation.

Therefore instead of instantiating with ChatGPT for the Inspiration Feedback module, we experiment with leveraging human heuristics.
The heuristics are ``if the inspiration titles are less related to the background, then more novel hypotheses are likely to be proposed.''.
With this heuristics-based past-feedback, MOOSE does perform better~(as shown in the tables in \S\ref{sec:experiment} and \S\ref{sec:analysis}).

This heuristics-based feedback is possible to be obtained by a language model since it has access to the novelty feedback of each hypothesis as well as the inspiration titles the hypothesis leveraged.
Here our contribution is to propose a useful framework for the TOMATO task, which is not limited by any LLMs for any module in the framework.
In the future, it is possible for more powerful LLMs to find better inspiration feedback than human heuristics.

\subsection{Qualitative Analysis on Hypotheses Generated From Different Methods}
\label{appen:qualitative_analysis}

We analyze four hypotheses from the baseline, MOOSE-base, MOOSE-base w/ future-feedback, and MOOSE-base w/ future and past-feedback~(MOOSE), where the four methods use the same passage to extract background.

\begin{itemize}

    \item Hypothesis from the baseline: \textit{Companies that prioritize customer understanding will have higher profitability than companies that do not prioritize customer understanding.}~(Evaluated by the expert, Validness: 4; Novelty: 1.5; Helpfulness: 2)
    \item Hypothesis from MOOSE-base: \textit{The level of empathy displayed by leaders in a startup environment influences employees' job satisfaction and organizational success through the mediation of employees' perceived likelihood of negative outcomes and expectation of enjoyment, as well as their propensity toward risky choices.}~(Evaluated by the expert, Validness: 3.5; Novelty: 3; Helpfulness: 3)
    \item Hypothesis from MOOSE-base w/ future-feedback: \textit{Female CMOs in startups, leveraging their higher levels of empathy, are more likely to prioritize customer satisfaction by actively listening to customer feedback, incorporating customer insights into decision-making processes, and providing personalized customer experiences. This employee prioritization of customer satisfaction is positively associated with higher levels of customer engagement, increased customer loyalty, and improved brand recall, as measured by objective metrics such as sales figures, customer retention rates, and brand recognition in the market. The influence of female CMOs on employee behavior is mediated by their ability to foster a caring relationship with customers, as supported by empirical data and statistical analysis.}~(Evaluated by the expert, Validness: 3.5; Novelty: 3.5; Helpfulness: 3.5)
    \item Hypothesis from MOOSE-base w/ future and past-feedback~(MOOSE): \textit{Female CMOs' empathy advantage influences their consideration of negative consequences of CSR initiatives, specifically in terms of employee well-being and job security. This relationship is moderated by individual differences in emotional intelligence. Additionally, the organizational culture and industry context will further influence the relationship between empathy advantage and consideration of negative consequences. The hypothesis will investigate whether female CMOs with higher levels of empathy are more likely to prioritize employee well-being and job security in the implementation of CSR initiatives, and whether this relationship is stronger in industries with a stronger emphasis on employee well-being and job security. It will also explore the mediating role of organizational culture and the moderating role of emotional intelligence in shaping the relationship between empathy advantage and consideration of negative consequences.}~(Evaluated by the expert, Validness: 4.5; Novelty: 4; Helpfulness: 4)
\end{itemize}

Analysis from the expert:

\begin{itemize}
    \item \textit{H1 falls short of challenging established assumptions or introducing a novel perspective beyond the widely accepted link between customer understanding and profitability.}
    \item \textit{Both H2 \& H3 center around a specific scenario involving female CMOs in startups and delve into their influence on customer satisfaction, employee behavior, and overall business results. From a research standpoint, this more focused approach points to a potential gap in the existing body of knowledge. Moreover, these two hypotheses surpass conventional understanding by considering how the empathy of female CMOs impacts employee behavior and business outcomes. They put forth a fresh viewpoint, suggesting that cultivating a compassionate rapport with customers, fostered by female CMOs, could positively affect customer engagement, loyalty, and brand recognition. These two hypotheses zoom in on a more specific context, introduce an innovative perspective, and probe a potential void in current research. They are anchored in the dynamic world of innovative business settings and propose a more nuanced and all-encompassing connection between variables.}
    \item \textit{H4 retains its relevance within a modern business landscape by scrutinizing the intersection of empathy, CSR initiatives, and the dynamics of organizations. This syncs seamlessly with the criterion of being rooted in an innovative business environment. Moreover, it shakes up established assumptions by considering the potential adverse outcomes of CSR initiatives and the role empathy plays in shaping decision-making within this context. This hypothesis delves into a more intricate and thorough exploration, examining a broader spectrum of factors and interactions within a specific context. Additionally, it imparts a deeper comprehension of the interplay between empathy, business choices, and organizational results. It grapples with a more complex and distinctive scenario, unearths possible gaps in the existing literature, and introduces a new angle on the role of empathy in the realm of business decisions.}
\end{itemize}

\subsection{Qualitative Analysis on Two MOOSE-Generated Hypotheses With High Expert Evaluation Scores}
\label{appen:old_two_qualitative_analysis}
In the following two grey boxes are two generated hypotheses from MOOSE with high expert evaluation scores~(appended to each hypothesis).
The expert's assessment of the two hypotheses is:




\mybox{gray!10}{\underline{Hypothesis~1:} \textit{The level of personalization in crowdfunding campaign storytelling, the influence of social media influencers who align with the campaign, the presence of trust indicators, and the emotional appeal of the campaign will positively impact potential donors' likelihood of making a donation. Additionally, the timing of donation requests and the type of social media influencers (e.g., celebrities vs. micro-influencers) will moderate this relationship. The perceived risk associated with the crowdfunding campaign will negatively moderate the relationship between the emotional appeal and donation likelihood.}~(Validness:~4.5; Novelty:~4.5; Helpfulness:~4.5)}

\mybox{gray!10}{\underline{Hypothesis~2:} \textit{Limited financial resources and limited access to networks and markets of women entrepreneurs in the manufacturing sector in developing countries may negatively impact their investment in corporate social responsibility (CSR) initiatives that promote gender equality in host countries. This relationship is further influenced by the intersectionality of gender and race, with women of color facing additional challenges. Additionally, the hypothesis considers the role of institutional factors, such as legal frameworks and policies, and the influence of patriarchal structures on women entrepreneurs' ability to invest in CSR initiatives.}~(Validness:~3.5; Novelty:~4.0; Helpfulness:~4.0)}

\textit{These two hypotheses both present a comprehensive view of the research narrative. It encompasses multiple hypotheses, including the primary one, as well as the mediation effect, which serves to elucidate the causal connection between the independent and dependent variables. Concurrently, both hypotheses outline the range of the effect --- namely, the circumstances in which this effect is applicable, under which scenarios where it might be weakened, and under which situation it could potentially be inverted.} 

\textit{In terms of novelty: 1.	Limited prior research or a gap in the existing literature. This means that there is a dearth of studies or information available on the subject, making it an unexplored area.
2.	Based on a new business setting. It is grounded in an innovative business environment, characterized by novel technologies, contemporary themes, and evolving business requirements.
3.	The topic offers a fresh and unique perspective that goes beyond conventional understanding. It might challenge existing assumptions, propose new theories, or present an unconventional approach.}

\subsection{Essential Factors for Good Social Science (and Business) Hypotheses}
\label{appen:factors_for_good_hypotheses}

According to business PhD students, counter-intuitive and novel hypotheses are the mostly favoured~(by top business journals).
Intuitive and novel hypotheses are also good but not as good as the counter-intuitive ones.
Here ``novel'' refers to ``not pointed out by existing literatures''.

Empirically they think of all the hypotheses on top business journals, around 20\% are counter-intuitive, leaving the remaining 80\% intuitive.

Counter-intuitive hypotheses tend to receive a lower validness evaluation compared to intuitive ones. 
For this reason, we highlight the counter-intuitive hypothesis in \S\ref{sec:qualitative_analysis}, even if it receives a lower score in validness than hypotheses in \S\ref{appen:old_two_qualitative_analysis}.

\subsection{An Example of Hypothesis Formulation via MOOSE}
\label{appen:an_example_hypothesis_formulation}

Here we show a complete flow of hypothesis discovery, by giving the output of relevant modules.
Specifically, the found background, found inspirations, generated suggestions, generated hypothesis, reality feedback, novelty feedback, and clarity feedback are copied in this section.

\begin{itemize}

\item Found background: 

\textit{Blind boxes, in terms of how they reach the consumers, are an innovation. Digging deeper into the issue, you will learn the very marketing of blind boxes is called probabilistic selling. That is to say, even after making the payment for a blind box, the buyer still has no idea what will be the contents inside. Suspense. A sense of thrill amid an air of expectation ensue. The sheer anticipation of a surprise can be intoxicating even as it can keep the consumer on tenterhooks, in a nice sort of way.}

\item Found inspirations: 

\textit{Inspiration 0:  "Surprise is like crack for your brain. Scientists at Emory and Baylor used MRIs to measure changes in human brain activity in response to a sequence of pleasurable stimuli, using fruit juice and water. The patterns of juice and water squirts were either predictable or completely unpredictable. Contrary to the researchers' expectations, the reward pathways in the brain responded most strongly to the unpredictable sequence of squirts."}

\textit{Inspiration 1:  "According to Tmall, China's largest B2C online marketplace, about 200,000 post-95 consumers bought blind boxes from its platform in 2019, with each spending CNY 20,000 per year on average; some fanatics even shelled out CNY 1 million on this "game of chances" to get the coveted limited-edition toys."}

\textit{Inspiration 2:  "Product differentiation is fundamentally a marketing strategy to encourage the consumer to choose one brand or product over another in a crowded field of competitors."}

\textit{Inspiration 3:  "The pandemic has shown how a lack of solid statistics can be dangerous. But even with the firmest of evidence, we often end up ignoring the facts we don't like"}

\textit{Inspiration 4:  "User-generated content (also known as UGC or consumer-generated content) is original, brand-specific content created by customers and published on social media or other channels."}

\textit{Inspiration 5:  "Higher Star Ratings Lead to More Orders.More stars really do lead to more sales.The data shows that products with higher star reviews have a much higher volume of purchases."}

\item Suggestions from Hypothesis Suggestor Module:

\textit{Suggestion 1:
Combining the background information on blind boxes and probabilistic selling with Inspiration 1 (surprise as a powerful reward stimulus) and Inspiration 2 (high spending on blind boxes by young consumers), a novel business research hypothesis could be: The element of surprise and unpredictability in probabilistic selling, such as blind boxes, triggers stronger reward responses in the brain, leading to higher consumer spending and loyalty, particularly among younger demographics. This hypothesis suggests that the psychological impact of surprise can be a significant driver of consumer behavior and purchasing decisions, especially for novel product categories.}

\textit{Suggestion 2:
Integrating the background context with Inspiration 3 (product differentiation strategy) and Inspiration 5 (user-generated content), another innovative hypothesis could be: Encouraging user-generated content around the surprise and reveal aspects of blind box products can serve as a powerful product differentiation strategy, setting brands apart in a competitive market. By leveraging the excitement and social sharing around the unboxing experience, businesses can create a unique brand identity and foster a sense of community among their customers.}

\textit{Suggestion 3:
Drawing from the background information, Inspiration 4 (the importance and challenges of evidence-based decision making), and Inspiration 6 (the impact of star ratings on sales), a third hypothesis could be: Despite the proven effectiveness of probabilistic selling in driving consumer engagement and sales, businesses may be reluctant to adopt this strategy due to a lack of traditional, evidence-based metrics and the perceived risks associated with unpredictable outcomes. This hypothesis highlights the potential tension between data-driven decision making and the benefits of innovative, surprise-based selling strategies.}

\item Hypothesis from Hypothesis Proposer Module: 

\textit{User-generated content, such as unboxing videos and reviews, can significantly influence the purchasing decisions of potential blind box buyers, particularly among younger demographics. The excitement and social sharing around the unboxing experience, as depicted in user-generated content, can amplify the psychological reward of surprise and drive higher sales, making it an effective tool for businesses to differentiate their products and increase consumer engagement in the blind box market.}

\item Reality Feedback:

\textit{The given hypothesis about the influence of user-generated content on blind box purchasing decisions among younger demographics seems to reflect reality to a certain extent. However, there are a few points to consider that might suggest the hypothesis does not entirely capture the reality:}

\textit{1. Generalization: The hypothesis assumes that user-generated content significantly influences all younger demographics. In reality, the impact may vary depending on factors such as personal preferences, cultural background, and socio-economic status.}

\textit{2. Other influential factors: While user-generated content can be an important factor in purchasing decisions, it may not be the sole or most significant driver. Other aspects like product quality, price, brand reputation, and peer influence might play equally important or even more crucial roles.}

\textit{3. Limited scope: The hypothesis focuses on the excitement and social sharing aspect of unboxing experiences. However, it does not consider the potential negative impact of user-generated content, such as disappointing unboxing experiences or negative reviews that could discourage purchases.}

\textit{4. Assumption of psychological reward: The hypothesis assumes that the surprise element in blind box purchases always leads to a positive psychological reward. In reality, some consumers might experience disappointment or frustration if they do not receive the desired item, which could deter future purchases.}

\textit{5. Lack of empirical evidence: The hypothesis presents a plausible explanation but does not provide concrete empirical evidence}

\item Novelty Feedback:

\textit{Based on the provided inspirations and the possibly related business literature paragraph, the research hypothesis appears to be novel and not a direct semantic copy of any of the given information. The hypothesis combines insights from various sources, such as the psychological impact of surprise, the popularity of blind boxes among younger demographics, the role of product differentiation, and the influence of user-generated content on purchasing decisions. While the hypothesis draws from these inspirations, it synthesizes them into a unique and specific context related to the blind box market and the impact of user-generated content on consumer engagement and sales. The provided literature paragraph discusses language modeling and large language models, which is not directly related to the hypothesis.To further enhance the novelty of the hypothesis, you could consider the following suggestions:}

\textit{1. Explore the potential long-term effects of user-generated content on brand loyalty and customer retention in the blind box market.}

\textit{2. Investigate the role of social media platforms and influencers in amplifying the impact of user-generated content on blind box sales.}

\textit{3. Examine the potential differences in the effectiveness of user-generated content across various product categories within the blind box market (e.g., collectible toys, beauty products, or snacks).}

\textit{4. Analyze the impact of user-generated content on the perceived value and collectibility of blind box items among consumers.}

\item Clarity Feedback:

\textit{The given hypothesis is relatively clear and specific, as it identifies the relationship between user-generated content and purchasing decisions of potential blind box buyers. It also specifies the target demographic (younger buyers), the psychological factors at play (excitement, surprise, and social sharing), and the potential benefits for businesses (product differentiation and increased consumer engagement). However, there is still room for improvement in terms of specificity.To make the hypothesis more specific, consider the following:}

\textit{1. Quantify the extent of the influence of user-generated content on purchasing decisions (e.g., "can increase the likelihood of purchase by X").}

\textit{2. Specify the age range of the "younger demographics" (e.g., "particularly among buyers aged 18-34").}

\textit{3. Clarify which types of user-generated content have the most significant impact (e.g., "especially unboxing videos that showcase the excitement and surprise of the experience").}

\textit{4. Indicate the platforms or channels where user-generated content is most effective (e.g., "on social media platforms such as Instagram and TikTok").}

\textit{5. Provide a more detailed explanation of how user-generated content amplifies the psychological reward (e.g., "by creating a sense of anticipation and vicarious experience for potential buyers").6. Quantify the potential}

\end{itemize}

\FloatBarrier

\subsection{Future Directions}
\label{appen:future_directions}

This work discovered the possibility of LLMs to propose novel research hypotheses. But it mainly focuses on the social science and business disciplines. It would be very interesting to investigate how LLMs can induce novel hypotheses for other disciplines~(especially nature science domains).

In addition, the MOOSE framework could be further improved to induce more valid and novel hypotheses for social science and business domains.

From the aspect of human-AI interaction, it would be also interesting to see how MOOSE can act as an AI Copilot to assist scientists in hypothesis discovery.




\begin{algorithm}[tb]
\caption{Algorithm for MOOSE}
\label{alg:algorithm}
\textbf{Input}: Raw web corpus $C$, related surveys $S$\\
\textbf{Parameter}: Total iterations for \emph{past-feedback} $M$, total iterations for \emph{present-feedback} $N$\\
\textbf{Output}: A list of hypotheses $H$ 

\begin{algorithmic}[1] 
\FOR{$c$ in $C$}
\STATE $b$, $b\_reason$ = Background\_Finder($c$)
\IF{$b$ $==$ None}
\STATE continue
\ENDIF
\FOR{iteration k $\in$ 0...$M$}
\IF{$k$ $!=$ 0}
\STATE $past\_f$ = Inspiration\_Feedback($t$, $h$, $present\_f$)
\ELSE
\STATE $past\_f$ = None
\ENDIF
\STATE $t$, $t\_reason$ = Inspiration\_Title\_Finder($C$, $b$, $b\_reason$, $past\_f$)
\STATE $p$ = find\_passage\_by\_title($t$, $C$)
\STATE $i$ = Inspiration\_Finder($b$, $b\_reason$, $p$, $t\_reason$)
\STATE $s$ = Hypothesis\_Suggestor($b$, $i$)
\STATE $h$ = Hypothesis\_Proposer($b$, $i$, $s$)
\FOR{iteration t $\in$ 0...$N$}
\STATE $cf$, $rf$, $nf$ = Clarity\_Checker($h$), Reality\_Checker($h$), Novelty\_Checker($h$, $S$)
\STATE $present\_f$ = [$cf$, $rf$, $nf$]
\STATE $h$ = Hypothesis\_Proposer($b$, $i$, $s$, $h$, $present\_f$)
\ENDFOR
\STATE $H$.append($h$)
\ENDFOR
\ENDFOR
\STATE \textbf{return} $H$
\end{algorithmic}
\end{algorithm}
\subsection{Full Algorithm of the MOOSE Framework}
\label{appen:algo}
Algorithm~\ref{alg:algorithm} shows the full algorithm of the proposed framework.

\end{document}